\documentclass{article} % For LaTeX2e
\usepackage{iclr2021_conference,times}
\usepackage[english]{babel}
\usepackage{amsmath,amsfonts,bm}

\def\eqref#1{equation~\ref{#1}}
\def\1{\bm{1}}

\def\vp{{\bm{p}}}

\DeclareMathAlphabet{\mathsfit}{\encodingdefault}{\sfdefault}{m}{sl}
\SetMathAlphabet{\mathsfit}{bold}{\encodingdefault}{\sfdefault}{bx}{n}
\newcommand{\tens}[1]{\bm{\mathsfit{#1}}}

\def\tE{{\tens{E}}}

\newcommand{\R}{\mathbb{R}}

\usepackage{graphicx}
\usepackage{wrapfig}
\usepackage{subfig}
\usepackage{float}
\usepackage{subfiles}
\iclrfinalcopy
\title{Data-Driven Shadowgraph Simulation of a 3D Object}
\author{Anna Willmann$^1$, Patrick Stiller$^1$, Alexander Debus$^1$, Arie Irman$^1$, \\ \textbf{Richard Pausch}$^1$,
\textbf{Yen-Yu Chang}$^1$, \textbf{Michael Bussmann}$^{1,2}$, \textbf{Nico Hoffmann}$^1$\\
\\
$^1$ Helmholtz-Zentrum Dresden - Rossendorf, Bautzner Landstrasse 400, 01328 Dresden, Germany\\
$^2$ CASUS - Center for Advanced Systems Understanding, Untermarkt 20, 02826 Görlitz, Germany
}

\usepackage{scalerel,stackengine}
\stackMath
\newcommand\reallywidehat[1]{%
\savestack{\tmpbox}{\stretchto{%
  \scaleto{%
    \scalerel*[\widthof{\ensuremath{#1}}]{\kern-.6pt\bigwedge\kern-.6pt}%
    {\rule[-\textheight/2]{1ex}{\textheight}}%WIDTH-LIMITED BIG WEDGE
  }{\textheight}% 
}{0.5ex}}%
\stackon[1pt]{#1}{\tmpbox}%
}

\begin{document}

\maketitle

\begin{abstract}
In this work we propose a deep neural network based surrogate model for a plasma shadowgraph - a technique for visualization of perturbations in a transparent medium.
We are substituting the numerical code by a computationally cheaper projection based surrogate model that is able to approximate the electric fields at a given time without computing all preceding electric fields as required by numerical methods. This means that the projection based surrogate model allows to recover the solution of the governing 3D partial differential equation, 3D wave equation, at any point of a given compute domain and configuration without the need to run a full simulation.
This model has shown a good quality of reconstruction in a problem of interpolation of data within a narrow range of simulation parameters and can be used for input data of large size.
\end{abstract}

\section{Introduction}
Simulations of physical processes are required in theory development and give better understanding of complex dynamics that are involved into phenomena. The more complicate system is considered the more compounded is the model of it and higher requirements to the computational power appear. For simplification of such models there are used surrogate models - models, that are created to study only certain aspects of processes that an original model represents precisely but in the same time other dynamics are excluded or have not expected behavior. Surrogate models can reduce time consumption of a research with an acceptable loss in accuracy of results.

Plasma shadowgraph is one of diagnostic techniques that provides a visualization of perturbations in a transparent medium as such phenomena are not visible by human eyes. This technique is based on refraction of probe rays when they are distributing through a medium, focused and specifically filtered in order to represent differently refracted rays by light and dark zones\citep{shadowg}.

Application of a plasma shadowgraph to some phenomena in plasma such as for example laser wakefield acceleration\citep{wakefield} can be not trivial due to intense interaction between particles inside plasma and the micron scale of fluctuations. For a correct analysis of the experimental data we need high quality simulations of these processes.

Simulation of a plasma shadowgraph consists of two steps.
At first we need to approximate the solution of Maxwell's equations and then we can calculate propagation of light in free space from Fourier optics. The model that is proposed in this paper is supposed to approximate the solution on the first step of the shadowgraph simulation, for simplification of the problem, we consider only the electric field component.

The main contribution of this work is a data-driven reduced-order model for approximation of the numerical simulation of large 3D computational domains. The neural network is approximating the solution for a simplified version of Maxwell's equations in a context of the electric field propagation problem, that can be described by the following equation:

\begin{equation*}\label{eq:ewave}
    \displaystyle \frac{\partial^2\displaystyle \tE}{\partial t^2} - \displaystyle \frac{1}{\mu\epsilon}\displaystyle\nabla^2\displaystyle \tE = 0
\end{equation*}

where $\displaystyle \tE$ is the electric field, $\displaystyle \mu$ and $\displaystyle \epsilon$ are permeability and permittivity of a medium.

The model is capable of reconstructing new simulations of the electric field for different parameters within a range limited by parameter values of existed simulations. Finally, we will be analysing the applicability of our approach for interpolation as well as extrapolation in parameter space compared to ground-truth data. 

\section{Related works}
Artificial neural network based models are widely applied in the field of radiophysics and in particular for approximation of solution for Maxwell's equations. Thus, for example physics informed neural networks, introduced in \citep{pinn} were applied in \citep{pinn_me_mmd} for approximation of solution of the frequency domain Maxwell's equation in the context of metamaterial design and for approximation of time-domain electromagnetic simulations derived by Maxwell's equations in \citep{pinn_me_tdes}. Another example of a surrogate model for Maxwell's equations is presented in the paper \citep{gm_em}, as opposed to previous two works, there authors use not only fully connected architecture but also convolutional architecture to archive better quality solution approximation of Maxwell’s equations over an arbitrary dielectric permittivity. In these works models approximate a solution directly based on the input parameters. Such method is intuitive but with increasing of outputs' size one can encounter a problem of high resource consumption. Another kind of methods, projection based models that reduce dimensionality of the original model and approximate solution in a reduced space. In \citep{pbm_me_tema} authors proposed an autoencoder based architecture for approximation of Maxwell's equations solution. A convolutional autoencoder decreases dimensionality of input data and evolution in time is recurrently computed by a modified LSTM\citep{lstm} network on the reduced space. Another work that proposes an autoencoder based solver for Maxwell's equations is \citep{ann_wp} for the problem of wave propagation. 

The model proposed in our work solves the problem of approximation using architecture of an autoencoder in order to reduce number of operations at approximation itself and in the same time replaces recurrent computations of each next time point by a direct mapping of parameters to a solution approximation in a reduced space. 

\section{Method}
The model consists of two parts, the first one, autoencoder, reduces dimensionality of input data and transforms it back to the original size and the second one, projection approximator, approximates the solution in a reduced space. The structure of the autoencoder is adopted from \citep{deepfluid}, where it is used as a part of a reduced order model for fluid dynamics simulations. The encoder consists of convolutional downsamling layers, each is preceded by a block of convolutional layers and an additive skip connection between them, the decoder is structured in the same way but layers are arranged in a reverse order. In addition, the last downsampling layer is followed by one more block of convolutional layers and afterwards there is applied a linear layer to reduce vector size to the desirable size of the latent space. In this work we define number of downsampling layers by the formula proposed in \citep{deepfluid} depending on a size of input, number of preceded convolutional layers is set to 4. Each layer in the network is followed by the activation function LeakyReLU\citep{lrelu} with leak of 0.2. 

The projection approximator can be seen as multi-layer perceptron architecture consisting of 4 fully connected layers: the input layer of size $\displaystyle k+1$, 2 hidden layers and the output layer of latent size $\displaystyle l$, each hidden layer is followed by $\displaystyle sin(x)$ activation function that captures the functional relationship of adjacent latent codes. 

\textbf{\\Compression of 3D simulation data:} let us denote the encoder by $\displaystyle R:  \displaystyle \R^{\displaystyle h\times \displaystyle w\times \displaystyle d} \rightarrow \displaystyle \R^{\displaystyle l}$, where $\displaystyle h$, $\displaystyle w$, $\displaystyle d$ - height, width and depth of an input volume $\displaystyle \tE^{\left (\displaystyle t\right )}$ at time point $\displaystyle t$, $\displaystyle l$ is a size of its latent representation. 
Then the decoder is denoted by $\displaystyle G: \displaystyle \R^{\displaystyle l} \rightarrow  \displaystyle \R^{\displaystyle h\times \displaystyle w\times \displaystyle d}$.
This autoencoder is trained by minimizing the supervised reconstruction loss $\mathcal{L}_{R,G}$ of an original volume $\displaystyle \tE^{\left (\displaystyle t\right )}$ and a reconstructed one $\reallywidehat{\displaystyle \tE}^{\left (\displaystyle t\right )} = G(R(\displaystyle \tE^{\left (\displaystyle t\right )}))$:
\begin{equation*}
    \displaystyle \mathcal{L}_{R,G}(\displaystyle \tE^{\displaystyle \left (t\right )})=\displaystyle ||\displaystyle \tE^{\left (\displaystyle t\right )}-\reallywidehat{\displaystyle \tE}{}^{\left (\displaystyle t\right )}||_1
\end{equation*}

\textbf{\\Projection approximation in Latent Space:} the projection approximator is learning the mapping \mbox{$\displaystyle F:\displaystyle \R^{\displaystyle k+1} \rightarrow \displaystyle \R^{\displaystyle l}$}, where $\displaystyle k$ is a number of simulation parameters and one additional parameter is a time point at which a solution is to be approximated. The objective function of this network is the supervised approximation error,
\begin{equation*}
    \displaystyle \mathcal{L}_F(R(\displaystyle \tE^{\left (\displaystyle t\right )}), \displaystyle \vp, \displaystyle t) = \displaystyle ||R(\displaystyle \tE^{\left (\displaystyle t\right )})-F(\displaystyle \vp,\displaystyle t)||_2
\end{equation*}

where $\displaystyle \vp$ is a vector of simulation parameters and $\displaystyle t$ some point in time. Pretrained autoencoder and projection approximator allows us to reconstruct the solution of our system for different points as of $\displaystyle \tE_{predicted}^{\left (\displaystyle t\right )}= \displaystyle G(\displaystyle F(\displaystyle \vp, \displaystyle t))$. 

\section{Results}

\begin{wrapfigure}[15]{r}{5.5cm}
    \begin{center}
    \vspace*{-7mm}
    \includegraphics[scale=0.32]{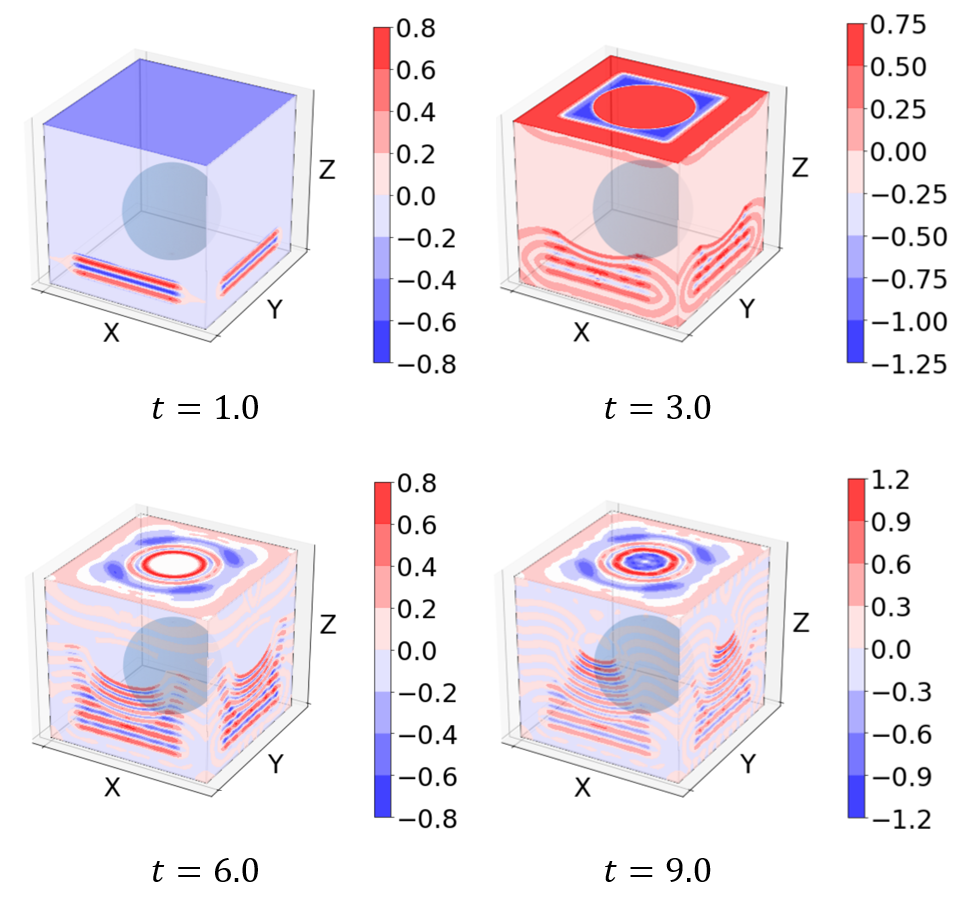}
    \caption{Example of simulation}
    \label{fig:cubesim}
    \end{center}
\end{wrapfigure}

For validation of the proposed model there are used simulations of beam propagation. 
Each simulation is an approximation of electric field propagation through a cell with a sphere in the middle that is defined by a radius $\displaystyle r$ and a refractive index $\displaystyle n$, calculated by finite-difference time-domain method\citep{ftdt} using a library Meep\citep{meep}. An example of such simulation is shown in figure \ref{fig:cubesim}. Surfaces of a cube represent middle slices of the cell from corresponding planes. 

The cell has physical size of 12$\displaystyle \mu m$ in each dimension with perfectly matched layer of 2$\displaystyle \mu m$ for the absorption at the boundary, proposed in \citep{pml}. Beam propagates in direction of axes Z, time points are given in units, one unit corresponds to $104.17 \mu s$. All parameters of simulations are given in supplementary material.

For training of the autoencoder such simulations were used with the following ranges of varying parameters: radius in $\displaystyle [2.0 \mu m, 4.0\mu m]$ with a step of 0.5$\displaystyle \mu m$ and refractive index in $\displaystyle [1.1, 1.7]$ with a step of 0.1. The same data in the reduced space is used for training of the projection approximator. Permutations of these values bring to 35 simulations that were used as training data for the autoencoder, one simulation approximates the electric field at each time point in range $\displaystyle [0.0, 10.0]$ with a step of 0.03125 time units, in total each simulation consists of 321 files where each file is a 3D array of size $193\times 193\times 193$ and takes 55Mb of memory and ca. 18Gb for a one full simulation.

\subsection{Training of model}
The networks were trained sequentially since the projection approximator requires a large number of epochs for convergence which would be computationally unfeasible in an end-to-end setting. Therefore, the autoencoder was trained first allowing us to precomputed the latent codes of all volumes of our training. The precomputed latent codes are then used for further training of the projection approximator. Parameters of networks were optimized by Adam optimizer\citep{adam} with learning rate of $0.0001$ over 160 iterations (autoencoder) and $0.001$ over 18400 iterations (projection approximator). Training of autoencoder was performed on 28 GPUs NVIDIA V100 for $29\;h$ and of projection approximator on 8 GPUs for $6\;h$.

\subsection{Analysis of latent code}

The analysis of temporal evolution of certain latent codes in the reduced space can give us better understanding and interpretation of the compressed representation.

\begin{wrapfigure}[9]{r}{5.5cm}
    \begin{center}
    \vspace*{-5mm}
    \includegraphics[scale=0.25]{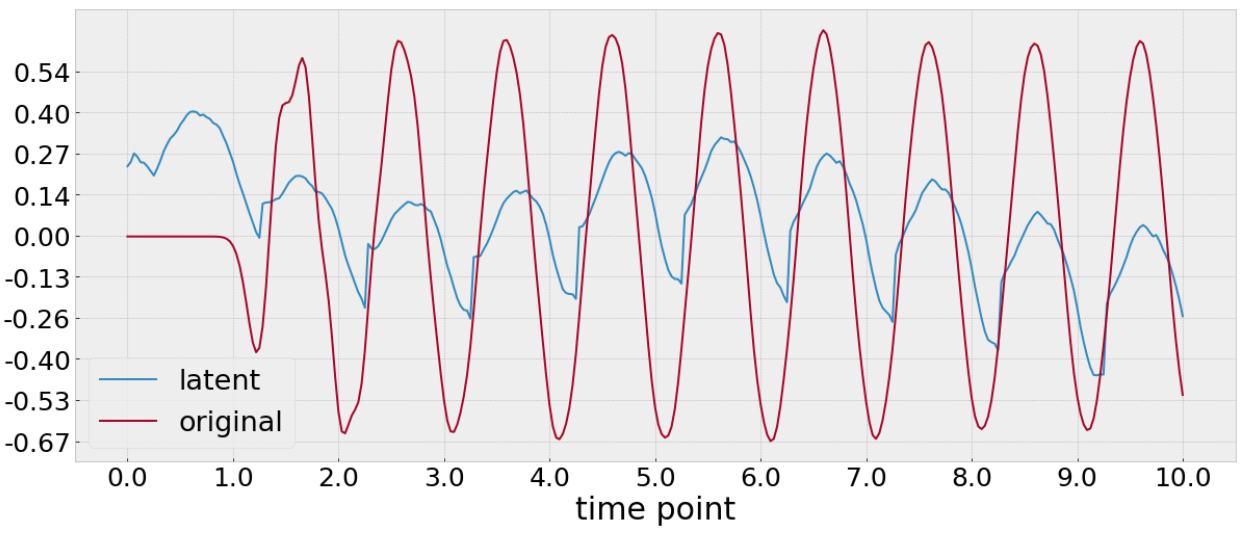}
    \caption{Evolution of components in the original and reduced spaces}
    \label{fig:latorig}
    \end{center}
\end{wrapfigure}
Figure \ref{fig:latorig} illustrates two series: one series shows the time evolution of a single voxel of our original input volume while the other series shows the time evolution of a certain latent dimension of the very same dataset.

There we see that the wave pattern with its period is preserved over encoding, then we can conclude that the field is compressed with maintaining certain physical dependencies from the original volume while parameters of simulation correspond to the range of values in latent representation.

\subsection{Generation of simulations}

The interpolation of simulations between known parameter values was performed successfully for the described model. 

\begin{figure}[h!]
  \centering
  \includegraphics[width=0.8\textwidth]{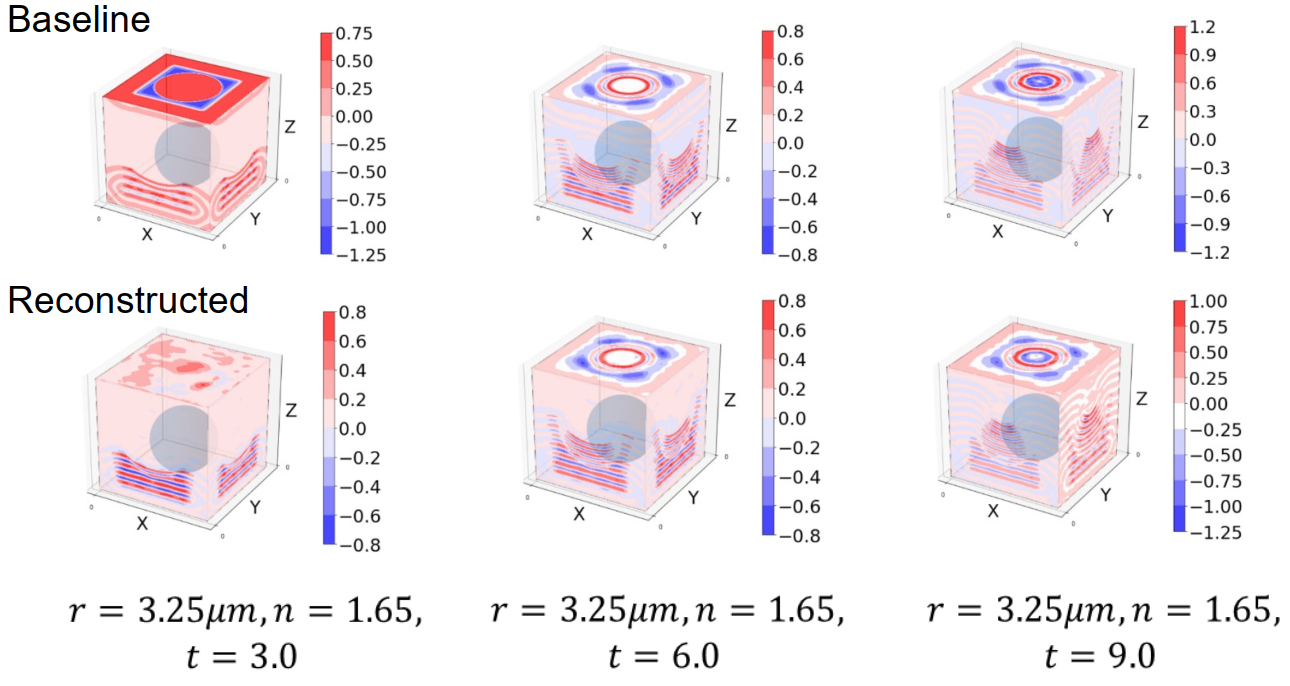}
  \caption{The reconstructed volumes clearly resemble the field data of the 3D wave equation.}
  \label{fig:rec}
\end{figure}

\begin{figure}[t]
  \centering
  \includegraphics[width=0.8\textwidth]{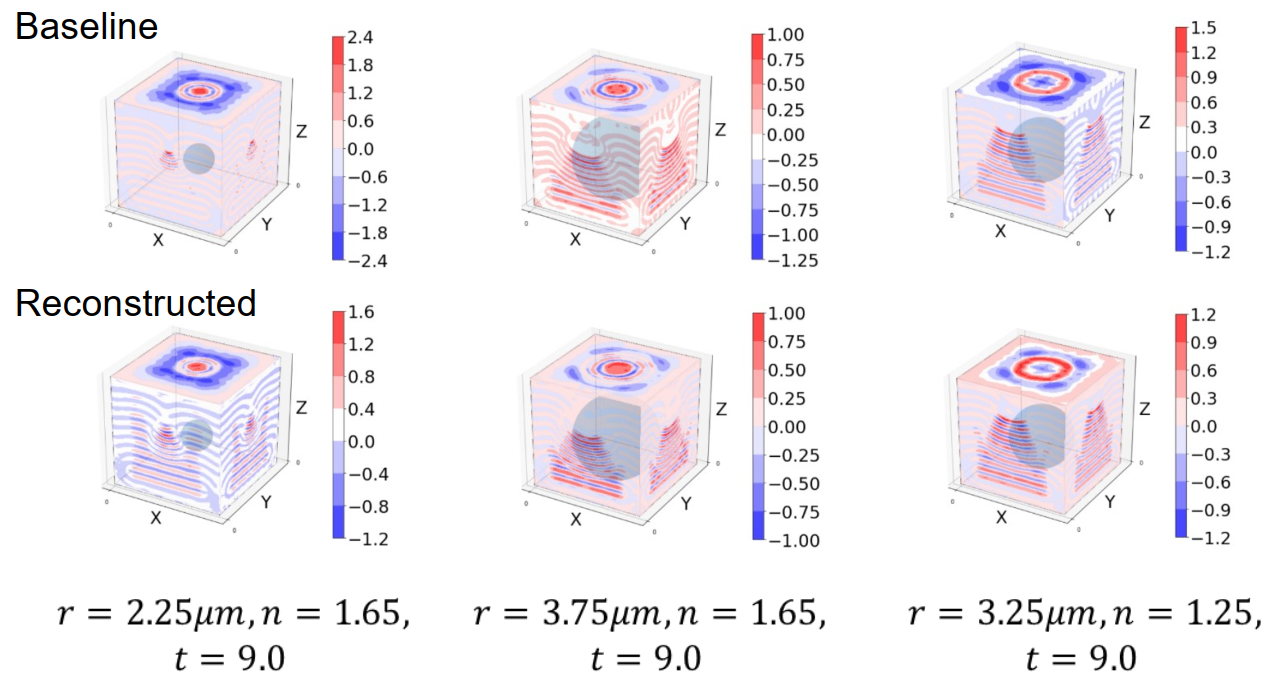}
  \caption{Examples of reconstruction}
  \label{fig:rec2}
\end{figure}

Figures \ref{fig:rec}, \ref{fig:rec2} show several examples of reconstruction. 
The model has shown an ability to reconstruct propagation of the field for all considered time points in the training set maintaining the structure of the field including refraction of the field in the location of a sphere as well as a distance on which the field is propagated until a certain point in time.

Examples from figure \ref{fig:rec} relate to time points 3.0, 6.0 and 9.0 respectively for a simulation with the same radius and refractive index. Examples from figure \ref{fig:rec2} show that the model is able to recognize different input radii of a sphere as well as different refractive indices respectively. Table \ref{tab:err_inter} provides average reconstruction errors over each simulation. 

\begin{wraptable}[9]{l}{6.5cm}
\vspace*{-4mm}
\caption{Reconstruction error for interpolation}
\label{tab:err_inter}
\begin{center}
\begin{tabular}{ll}
\multicolumn{1}{c}{\bf Parameters $\displaystyle r$, $\displaystyle n$ }  &\multicolumn{1}{c}{\bf $\displaystyle L_1$ error}
\\ \hline \\
$2.25\mu m$, $1.65$ &0.01639\\
 $3.25\mu m$, $1.25$ &0.01595\\
 $3.25\mu m$, $1.65$  &0.01813 \\
 $3.75\mu m$, $1.65$ &0.01823 \\
\end{tabular}
\end{center}
\end{wraptable}

The model can handle parameter values beyond the considered ranges (extrapolation) for training only within a small distance as new simulations contain features that did not appear in the training set. 

The time consumption for approximation of a single 3D volume is reduced by a factor of four compared to the computational time of conventional numerical method: the FTDT method implemented in Meep takes ca. $0.8\;s$ per time point while for the developed model it takes ca. $0.18\;s$. 
An even larger improvement in runtime can be gained as the projection approximator network not only allows us to recover certain subsequent time points but also enables us to fast-forward to any given time point of compute domain in $0.18\;s$.

\section{Conclusions}
The fast reconstruction of experimentally accessible diagnostics is a crucial task for understanding very complex systems such as Laser plasma accelerators. A general strategy for solving the involved ill-posed inverse problems requires the optimisation of numerical simulations which is computationally very demanding. We are tackling that issue by a projection-based surrogate model that successfully approximates the governing 3D field propagation with stable behavior for interpolation in parameter space. Interestingly, we found that the autoencoder part of our architecture preserves time-dependent physical properties while also encoding information about the parameters describing the system. The surrogate model promises significant acceleration compared to numerical methods by allowing direct access to the solutions of the governing equation at any point in time without the need of time-stepping schemes. Additionally, the surrogate model promises an speedup by factor of four comparing to the conventional FTDT method even for simple time-stepping.


\newpage

\begin{thebibliography}{15}
\providecommand{\natexlab}[1]{#1}
\providecommand{\url}[1]{\texttt{#1}}
\expandafter\ifx\csname urlstyle\endcsname\relax
  \providecommand{\doi}[1]{doi: #1}\else
  \providecommand{\doi}{doi: \begingroup \urlstyle{rm}\Url}\fi

\bibitem[Albert et~al.(2014)Albert, Thomas, Mangles, Banerjee, Corde, Flacco,
  Litos, Neely, Vieira, Najmudin, Bingham, Joshi, and Katsouleas]{wakefield}
Felicie Albert, Alec Thomas, Stuart Mangles, S~Banerjee, Sébastien Corde,
  A.~Flacco, Michael Litos, D~Neely, J.~Vieira, Zulfikar Najmudin, Robert
  Bingham, Chandrashekhar Joshi, and T~Katsouleas.
\newblock Laser wakefield accelerator based light sources: Potential
  applications and requirements.
\newblock \emph{Plasma Physics and Controlled Fusion}, 56:\penalty0 084015, 07
  2014.

\bibitem[Bartlett(2018)]{gm_em}
Ben Bartlett.
\newblock A "generative" model for computing electromagnetic field solutions.
\newblock \emph{Stanford CS229 Projects}, 2018\penalty0 (233), 2018.

\bibitem[Berenger(1994)]{pml}
Jean-Pierre Berenger.
\newblock A perfectly matched layer for the absorption of electromagnetic
  waves.
\newblock \emph{Journal of Computational Physics}, 114\penalty0 (2):\penalty0
  185 -- 200, 1994.
\newblock ISSN 0021-9991.

\bibitem[{Cheng} et~al.(2020){Cheng}, {Zhang}, and {Shao}]{ann_wp}
X.~{Cheng}, Z.~Y. {Zhang}, and W.~{Shao}.
\newblock A surrogate model based on artificial neural networks for wave
  propagation in uncertain media.
\newblock \emph{IEEE Access}, 8:\penalty0 218323--218330, 2020.

\bibitem[{Fang} \& {Zhan}(2020){Fang} and {Zhan}]{pinn_me_mmd}
Z.~{Fang} and J.~{Zhan}.
\newblock Deep physical informed neural networks for metamaterial design.
\newblock \emph{IEEE Access}, 8:\penalty0 24506--24513, 2020.

\bibitem[Hochreiter \& Schmidhuber(1997)Hochreiter and Schmidhuber]{lstm}
Sepp Hochreiter and Jürgen Schmidhuber.
\newblock Long short-term memory.
\newblock \emph{Neural Computation}, 9\penalty0 (8):\penalty0 1735--1780, 1997.

\bibitem[{Kane Yee}(1966)]{ftdt}
{Kane Yee}.
\newblock Numerical solution of initial boundary value problems involving
  maxwell's equations in isotropic media.
\newblock \emph{IEEE Transactions on Antennas and Propagation}, 14\penalty0
  (3):\penalty0 302--307, 1966.

\bibitem[Kim et~al.(2019)Kim, Azevedo, Thuerey, Kim, Gross, and
  Solenthaler]{deepfluid}
Byungsoo Kim, Vinicius Azevedo, Nils Thuerey, Theodore Kim, Markus Gross, and
  Barbara Solenthaler.
\newblock Deep fluids: A generative network for parameterized fluid
  simulations.
\newblock \emph{Computer Graphics Forum}, 38:\penalty0 59--70, 05 2019.

\bibitem[Kingma \& Ba(2017)Kingma and Ba]{adam}
Diederik~P. Kingma and Jimmy Ba.
\newblock Adam: A method for stochastic optimization.
\newblock \emph{arXiv preprint arXiv:1412.6980}, 2017.

\bibitem[Maas(2013)]{lrelu}
Andrew~L. Maas.
\newblock Rectifier nonlinearities improve neural network acoustic models.
\newblock 2013.

\bibitem[{Noakoasteen} et~al.(2020){Noakoasteen}, {Wang}, {Peng}, and
  {Christodoulou}]{pbm_me_tema}
O.~{Noakoasteen}, S.~{Wang}, Z.~{Peng}, and C.~{Christodoulou}.
\newblock Physics-informed deep neural networks for transient electromagnetic
  analysis.
\newblock \emph{IEEE Open Journal of Antennas and Propagation}, 1:\penalty0
  404--412, 2020.

\bibitem[Oskooi et~al.(2010)Oskooi, Roundy, Ibanescu, Bermel, Joannopoulos, and
  Johnson]{meep}
Ardavan~F. Oskooi, David Roundy, Mihai Ibanescu, Peter Bermel, J.D.
  Joannopoulos, and Steven~G. Johnson.
\newblock Meep: A flexible free-software package for electromagnetic
  simulations by the fdtd method.
\newblock \emph{Computer Physics Communications}, 181\penalty0 (3):\penalty0
  687 -- 702, 2010.
\newblock ISSN 0010-4655.

\bibitem[Raissi et~al.(2017)Raissi, Perdikaris, and Karniadakis]{pinn}
Maziar Raissi, Paris Perdikaris, and George~Em Karniadakis.
\newblock Physics informed deep learning (part i): Data-driven solutions of
  nonlinear partial differential equations.
\newblock \emph{arXiv preprint arXiv:1711.10561}, 2017.

\bibitem[Traldi et~al.(2018)Traldi, Boselli, Simoncelli, Stancampiano,
  Gherardi, Colombo, and Settles]{shadowg}
E.~Traldi, M.~Boselli, E.~Simoncelli, A.~Stancampiano, M.~Gherardi, V.~Colombo,
  and Gary~S. Settles.
\newblock Schlieren imaging: a powerful tool for atmospheric plasma diagnostic.
\newblock \emph{EPJ Techniques and Instrumentation}, 4:\penalty0 4, 05 2018.

\bibitem[{Zhang} et~al.(2020){Zhang}, {Hu}, {Jin}, {Deng}, {Wu}, and
  {Chen}]{pinn_me_tdes}
P.~{Zhang}, Y.~{Hu}, Y.~{Jin}, S.~{Deng}, X.~{Wu}, and J.~{Chen}.
\newblock A maxwell's equations based deep learning method for time domain
  electromagnetic simulations.
\newblock pp.\  1--4, 2020.

\end{thebibliography}
\end{document}

\section*{Supplementary material}

\subsection*{Parameters of simulations used for training data}\label{sec:params}

The cell has physical size of $12 \mu m$ in each dimension and resolution of simulation is $16 \text{ pixels per } \mu m$. 
Boundary condition is a perfectly matched layer for the absorption with width of $2 \mu m$ on each side of the cell. 
The source of the current starts to distribute the field in direction of axes Z from the starting point with coordinates $(-4,0,0)$ w.r.t. the center of the cell(coordinates of center are $(0,0,0)$), wavelength of the source is $1.0 \mu m$, size of the source is $(0, 8, 8)$, i.e. a flat source in space between absorbing layers. 
Center of a sphere is located in the middle of the cell - $(0,0,0)$, with defined radius and refractive index of material. Time points are given in units, one unit corresponds to $104.17 \mu s$.

\subsection*{Additional plots to latent space analysis}

\begin{figure}[h!]
    \begin{center}
    \includegraphics[scale=0.4]{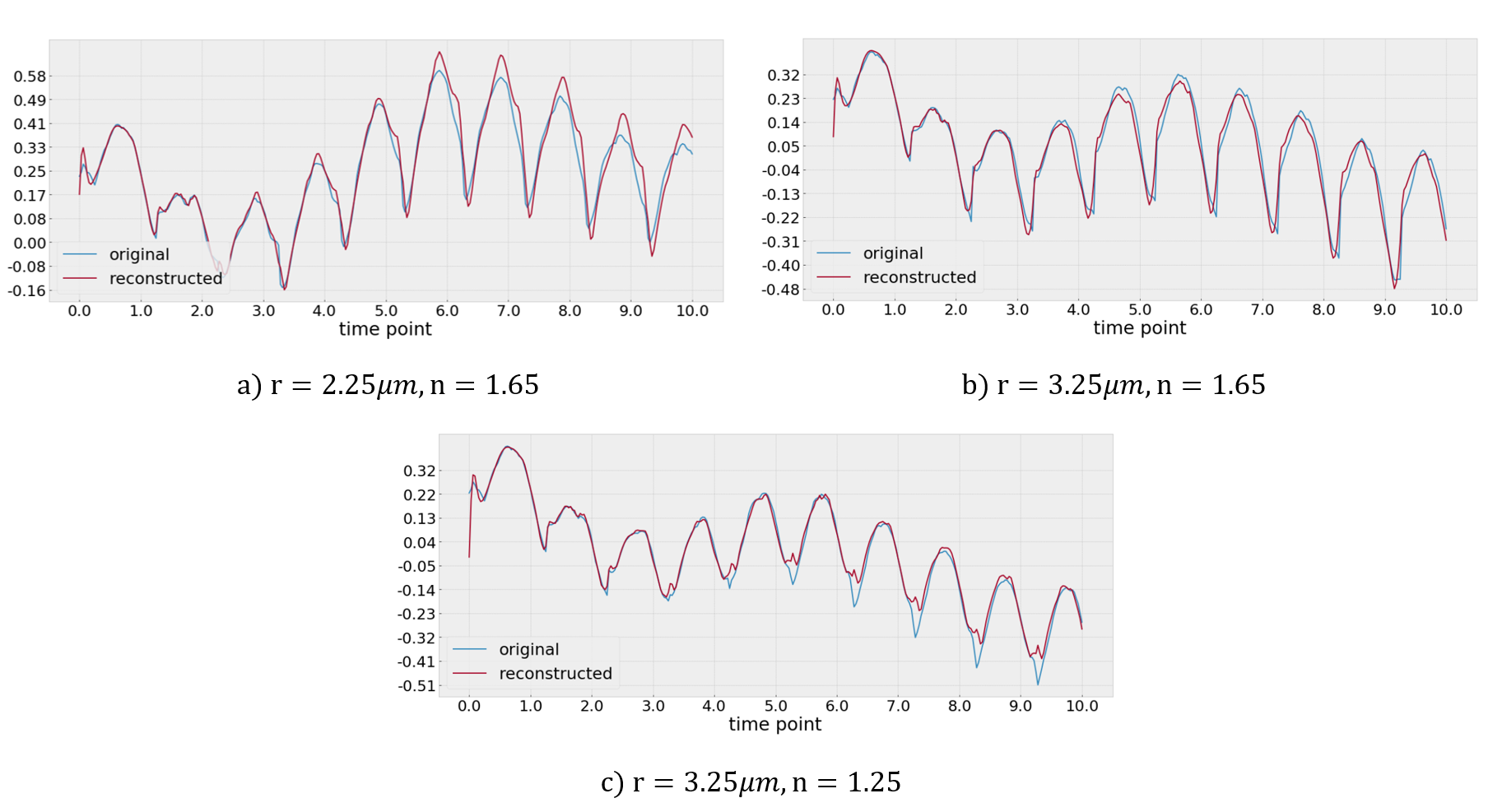}
    \caption{Evolution of a single component from simulations with different parameters in the reduced space}
    \label{fig:latan}
    \end{center}
\end{figure}

\subsection*{Additional results}
\begin{figure}[H]
    \begin{center}
    \includegraphics[width=0.65\textwidth]{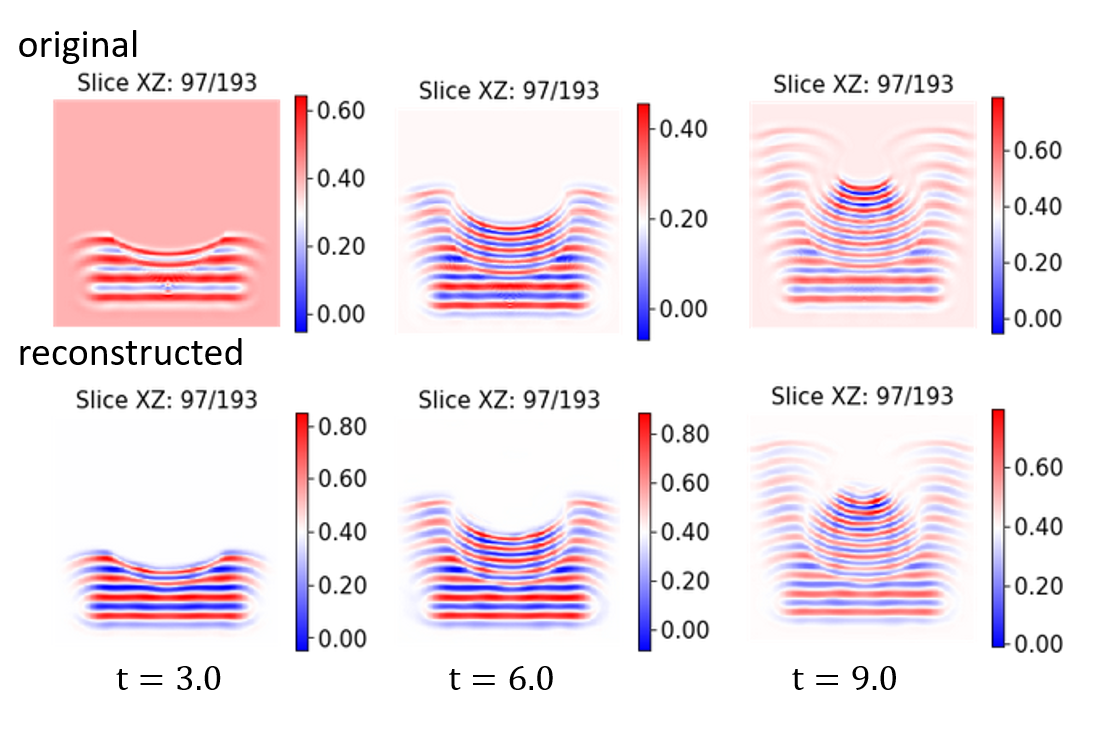}
    \caption{Reconstruction of simulation: $ r=3.25\mu m$, $ n = 1.65$}
    \label{fig:evol}
    \end{center}
\end{figure}

\begin{figure}[H]
    \begin{center}
    \includegraphics[width=0.8\textwidth]{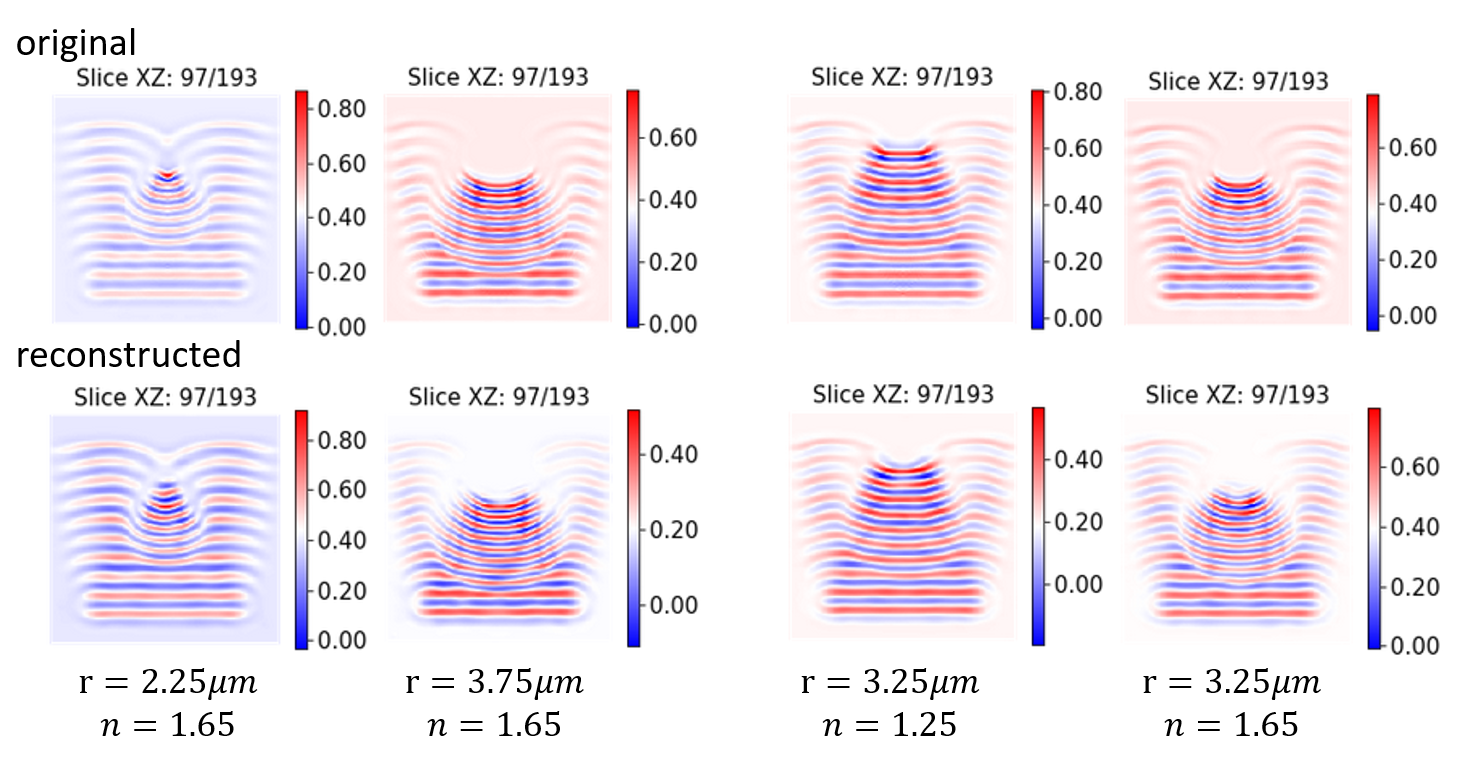}
    \caption{Examples of reconstruction: interpolation}
    \label{fig:interp}
    \end{center}
\end{figure}
\vspace*{-0.5cm}
\begin{figure}[H]
    \begin{center}
    \includegraphics[width=0.65\textwidth]{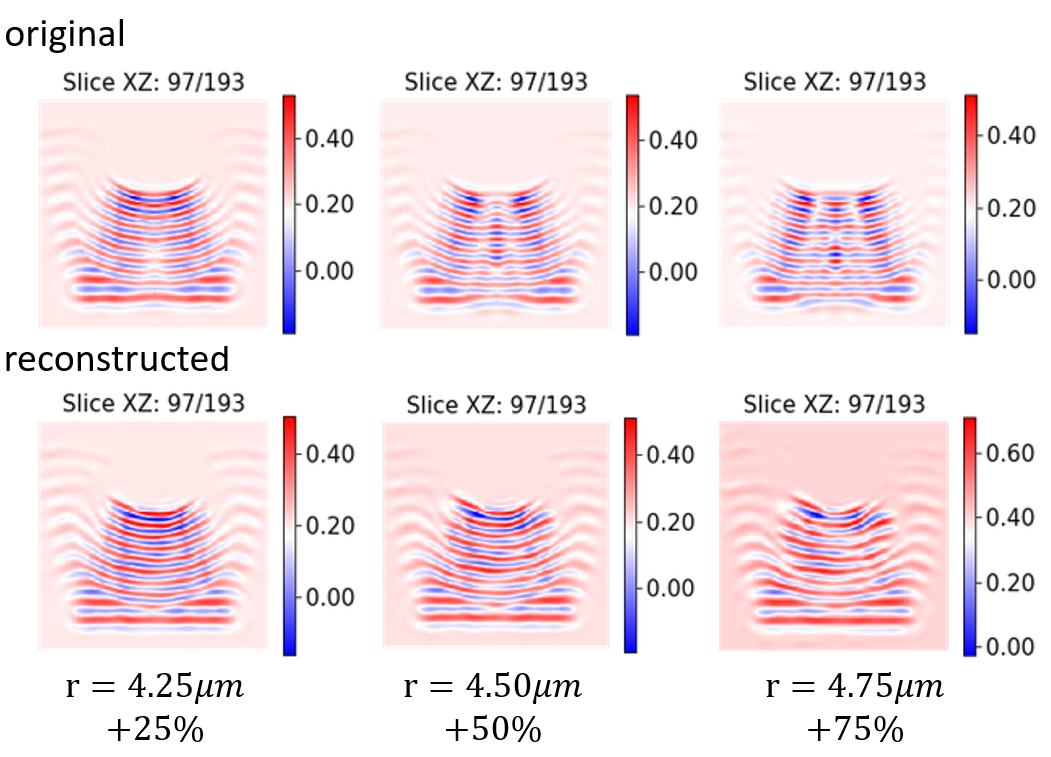}
    \caption{Examples of reconstruction: extrapolation of radius}
    \label{fig:extrp_rad}
    \end{center}
\end{figure}

\begin{figure}[H]
    \begin{center}
    \includegraphics[width=0.65\textwidth]{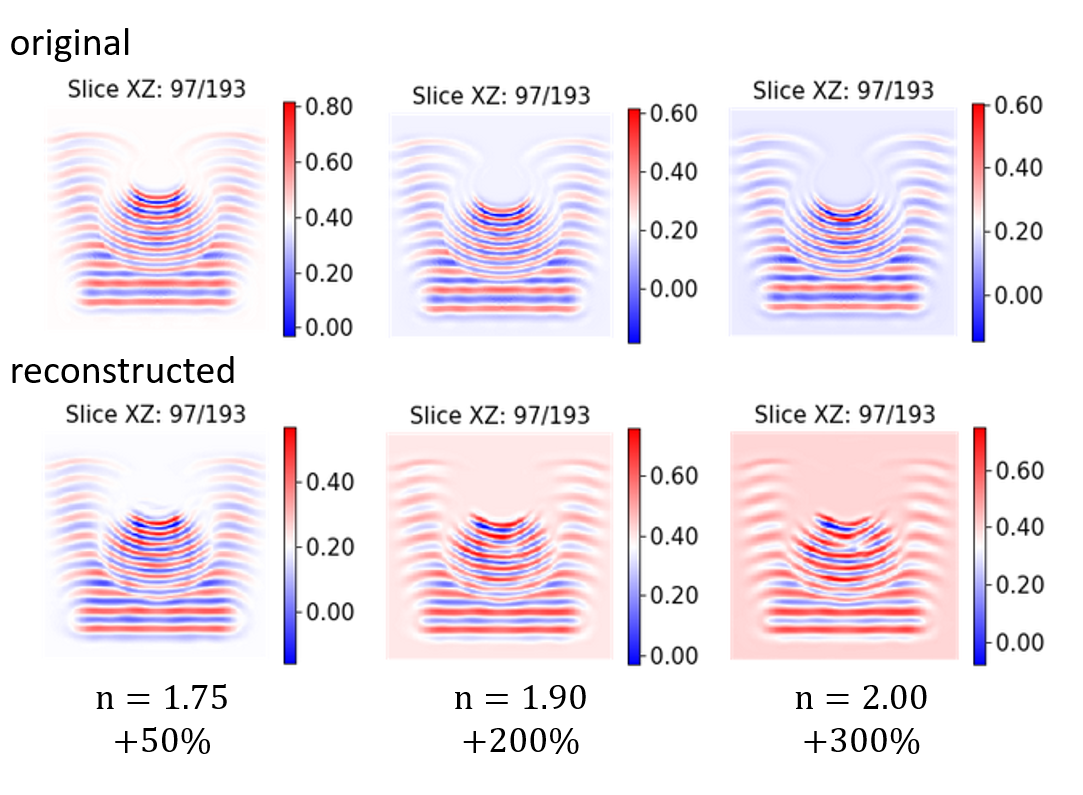}
    \caption{Examples of reconstruction: extrapolation of refractive index}
    \label{fig:extrp_ri}
    \end{center}
\end{figure}

\begin{table}[H]
\caption{Reconstruction error for extrapolation}
\label{tab:extr}
\begin{center}
\begin{tabular}{ll}
\multicolumn{1}{c}{\bf Parameters: $ r$, $ n$ }  &\multicolumn{1}{c}{\bf ($ L_1$) error}
\\ \hline \\
$4.25\mu m$, $1.65$ &0.01639\\
 $4.50\mu m$, $1.65$ &0.01826\\
 $4.75\mu m$, $1.65$  &0.01813 \\
 & \\
 $3.25\mu m$, $1.75$ &0.01877 \\
 $3.25\mu m$, $1.90$ &0.02161\\
 $3.25\mu m$, $2.00$  &0.02421\\
\end{tabular}
\end{center}
\end{table}